\documentclass{article}

\usepackage{jmlr2e}

\usepackage{graphicx} 
\usepackage{amsmath}
\usepackage{amssymb}
\usepackage{amsfonts}
\usepackage{algorithm}
\usepackage{algpseudocode}
\usepackage{enumitem}
\usepackage{appendix}
\usepackage{booktabs}
\usepackage{xcolor}

\usepackage{lastpage}
\jmlrheading{26}{2025}{1-\pageref{LastPage}}{11/4; Revised ?/??}{?/??}{??-????}{David Nabergoj and Erik Štrumbelj}
\ShortHeadings{Reducing NF complexity for MCMC preconditioning}{Nabergoj and Štrumbelj}
\firstpageno{1}

\begin{document}

\title{Reducing normalizing flow complexity for MCMC preconditioning}
\author{\name David Nabergoj \email david.nabergoj@fri.uni-lj.si \\
        \addr University of Ljubljana, Faculty of Computer and Information Science \\
        Ve\v{c}na pot 113, 1000 Ljubljana, Slovenia
        \AND
        \name Erik Štrumbelj \email erik.strumbelj@fri.uni-lj.si \\
        \addr University of Ljubljana, Faculty of Computer and Information Science \\
        Ve\v{c}na pot 113, 1000 Ljubljana, Slovenia
        \AND}

\editor{My editor}

\maketitle

\begin{abstract}
    Preconditioning is a key component of MCMC algorithms that improves sampling efficiency by facilitating exploration of geometrically complex target distributions through an invertible map.
    While linear preconditioners are often sufficient for moderately complex target distributions, recent work has explored nonlinear preconditioning with invertible neural networks as components of normalizing flows (NFs).
    However, empirical and theoretical studies show that overparameterized NF preconditioners can degrade sampling efficiency and fit quality. Moreover, existing NF-based approaches do not adapt their architectures to the target distribution.
    Related work outside of MCMC similarly finds that suitably parameterized NFs can achieve comparable or superior performance with substantially less training time or data.
    We propose a factorized preconditioning architecture that reduces NF complexity by combining a linear component with a conditional NF, improving adaptability to target geometry.
    The linear preconditioner is applied to dimensions that are approximately Gaussian, as estimated from warmup samples, while the conditional NF models more complex dimensions.
    Our method yields significantly better tail samples on two complex synthetic distributions and consistently better performance on a sparse logistic regression posterior across varying likelihood and prior strengths.
    It also achieves higher effective sample sizes on hierarchical Bayesian model posteriors with weak likelihoods and strong funnel geometries.
    This approach is particularly relevant for hierarchical Bayesian model analyses with limited data and could inform current theoretical and software strides in neural MCMC design.
\end{abstract}


\section{Introduction}
Markov Chain Monte Carlo (MCMC) is a family of numerical integration methods that are based on drawing samples from probability distributions.
The process involves starting with an initial state and evolving it using Markov kernel transitions, which yield a chain of states.
In many practical applications, each visited state is treated as a draw from the distribution of interest.
The quality of the obtained draws depends on the kernel's exploration efficiency.
Popular MCMC methods and software packages increase this efficiency by preconditioning the kernel.
Instead of directly sampling from the target distribution, the kernel is often applied in a linearly transformed space before transforming the samples back to the original space.
\cite{hoffman_neutralizing_2019} suggested that the transformation instead be performed with an invertible neural network as a component of a normalizing flow~\citep[NF,][]{papamakarios_normalizing_2021}.
This idea motivated many preconditioned MCMC samplers, including preconditioned Metropolis-Hastings~\citep{gabrie_efficient_2021, gabrie_adaptive_2022, brofos_adaptation_2022} and sequential samplers~\citep{karamanis_accelerating_2022, arbel_annealed_2021, matthews_continual_2022}.
The nonlinear NF preconditioner allows these methods to efficiently sample from distributions with nontrivial geometry, e.g., distributions that are not approximately Gaussian.

\cite{nabergoj_empirical_2024} recently performed an empirical analysis of NF architectures in MCMC and found that NFs with fewer parameters are preferable to those with many parameters.
This is reasonable, as several of their test distributions, including hierarchical Bayesian model (HBM) posteriors, are approximately Gaussian in many dimensions.
Such distributions can be suitably modeled by simpler NFs.
To adopt NFs as primary preconditioners in general-purpose MCMC packages, the considered architectures should efficiently model distributions with many approximately Gaussian dimensions.
At the same time, the architecture should have the capacity to model distributions with non-trivial geometry and preferably adapt to the degree of geometric complexity without requiring user input.
Unfortunately, the aforementioned samplers rely on NFs that require manual hyperparameter tuning, and~\cite{nabergoj_empirical_2024} similarly recommend such tuning, which can be time-consuming for the user.
Furthermore, these architectures do not distinguish between the geometrically simple and complex dimensions of the target distribution.
This represents a gap in the field and calls for an NF architecture whose parameterization adapts to the target complexity.

In this paper, we propose an NF architecture with reduced complexity by factoring it into a linear and a nonlinear transform according to MCMC draws.
The model linearly preconditions the approximately Gaussian subset of dimensions and nonlinearly preconditions the rest.
We select the approximately Gaussian subset according to a novel heuristic based on Wasserstein distance. 
Our model contains fewer parameters than the corresponding base NF and achieves better sampling efficiency on Bayesian model posteriors, as measured by kernelized Stein discrepancy (KSD).

\section{Related work}
The importance of nonlinear preconditioning in MCMC stems from the non-Gaussianity of various target distributions.
In HBMs,~\cite{neal_slice_2003} demonstrated the inefficiency of the Gibbs sampler on a synthetic funnel distribution.
\cite{betancourt_hamiltonian_2015} further discussed that such geometries are very prevalent in HBMs with sparse data and thus strong priors.
They suggest designing non-centered parameterizations~\citep{papaspiliopoulos_general_2007} to improve posterior geometries for HMC sampling.

Since then, many works suggested automated or learned model reparameterization~\citep{parno_transport_2018, gorinova_automatic_2020} with~\cite{hoffman_neutralizing_2019} suggesting nonlinear MCMC preconditioning based on NFs.
There have been several further advancements connecting NF preconditioning or global NF proposals with models in cosmology~\citep{gabrie_efficient_2021,karamanis_accelerating_2022, grumitt_sequential_2024}, field theory~\citep{gabrie_adaptive_2022}, molecular dynamics~\citep{gabrie_adaptive_2022}, chaotic systems~\citep{grumitt_flow_2024}, as well as HBMs and general sampling~\citep{arbel_annealed_2021, grumitt_deterministic_2022, matthews_continual_2022}.
The methods show improved sampling efficiency on target distributions with funnel geometries and multimodality, as well as ill-conditioned distributions.

There exists less systematic research regarding suitable NF architectures for MCMC preconditioning, though several recent works address different aspects of the problem.
In a recent empirical investigation,~\cite{nabergoj_empirical_2024} found that using NFs with $10^3$ to $10^4$ trainable parameters achieved better results than bigger models on targets with fewer than 1000 model parameters, indicating that overparametrization could be hindering the preconditioning performance of NFs.
\cite{grenioux_sampling_2023a} observed a reduction in sliced total variation and Kolmogorov-Smirnov distances between MCMC samples and the target distribution as the NF fit better matches the target distribution.
Outside of MCMC, several works proposed injecting domain knowledge into NFs via probabilistic graphical models~\citep[PGMs;][]{weilbach_structured_2020, wehenkel_graphical_2021, mouton_graphical_2022a}.
\cite{wehenkel_graphical_2021} found this approach to substantially outperform NF architectures without any injected knowledge on density estimation tasks.
Moreover, they found that such NFs only require a single layer to match the density estimation performance of architectures with multiple layers.
These ideas have also been explored in structured equation modeling and structured causal modeling~\citep{khemakhem_causal_2021, balgi_personalized_2022, javaloy_causal_2023a} with \cite{khemakhem_causal_2021} similarly reporting high efficiency at small sample sizes.
While inferring PGMs or causal graphs may not be directly applicable for arbitrary target distributions in MCMC, these works highlight the promise of designing an NF with a more explicit treatment of geometrically nontrivial dimensions, including a simplified treatment of approximately Gaussian dimensions.
Further,~\cite{wehenkel_graphical_2021} describe an approach for automatic determination of PGMs, and thus demonstrate that some degree of knowledge can be injected into NFs without user input.
This further motivates automatically identifying information that can inform preconditioner parameterization, such as approximate Gaussianity, as investigated in this work.

Our approach to tailoring preconditioner complexity to the target distribution can be related to Gibbs sampling~\citep{geman_stochastic_1984}.
A Gibbs step consists of updating a subset of state dimensions, then updating the remaining dimensions conditional on the former.
Our method similarly factorizes the preconditioner into a linear and a nonlinear component, each appropriately transforming it corresponding dimensions into an approximately Gaussian latent space.
However, rather than restricting ourselves to sampling with fixed dimension subsets as in Gibbs, we allow joint sampling of the entire distribution with an arbitrary MCMC sampler.
This allows MCMC to implicitly correct for suboptimal subset choices, while simplifying the geometry of each subset.

Funnel-shaped target distributions represent a relevant test case for our work, as successfully identifying approximately Gaussian dimensions can lead to a highly efficient reparameterization of the funnel geometry.
Several works already approached this problem, including slice sampling~\citep{neal_slice_2003}, which generalizes Gibbs updates by sampling under the multidimensional target density plot conditional on a slice variable; and Riemannian HMC~\citep{girolami_riemann_2011} which replaces the constant mass matrix (equivalent to preconditioning) with a position-specific metric based on a Riemannian manifold.
Recently,~\cite{gundersen_escaping_2025} proposed sampling from a generalized higher-dimensional HBM that reduces funnel sharpness, then numerically marginalizing out the parameters of the generalized model to recover samples from the target HBM.
The drawback of Riemannian HMC is the apparent numerical instability in the proposed Hamiltonian.
Slice sampling and generalized HBMs focus on the sampling strategy, but do not consider improving the underlying preconditioner.
Our direct focus in this work is reducing NF complexity for general MCMC preconditioning, potentially including non-HBM target distributions.

\section{Methods}
In this section, we introduce a heuristic for determining whether draws from a one-dimensional distribution are approximately Gaussian.
We use this heuristic to define a factorized NF preconditioner, which consists of a multivariate Gaussian model for such dimensions and a conditional NF for the remaining ones.

\subsection{On preconditioned MCMC}
We first state that a pushforward measure is invariant under its corresponding pushforward kernel. 
This justifies the use of preconditioners to construct valid Markov chains.

Let $(\mathcal{X}, \mathcal{F}), (\mathcal{Z}, \mathcal{G})$ be measurable spaces.
Let $K(x, \cdot)$ be a Markov kernel on $\mathcal{X}$, i.e., for each $x \in \mathcal{X}$, $K(x, \cdot)$ is a probability measure on $(\mathcal{X}, \mathcal{F})$, and for each measurable $B \subseteq \mathcal{X}$, the map $x \mapsto K(x, B)$ is measurable.
Let $f:\mathcal{X} \rightarrow \mathcal{Z}$ be a measurable bijection with a measurable inverse $f^{-1}: \mathcal{Z} \rightarrow \mathcal{X}$.
Define the pushforward kernel $\tilde{K}$ on $\mathcal{Z}$ as:
\begin{align}
    \tilde{K}(z, A) \triangleq K(f^{-1}(z), f^{-1}(A)) \textrm{ for all } z \in \mathcal{Z}, A \in \mathcal{G}.
\end{align}
For fixed $z$, $\tilde{K}(z, \cdot)$ is a probability measure because $K(f^{-1}(x), \cdot)$ is a probability measure and $f^{-1}$ preserves measurability.
For fixed $A$, the map $z \mapsto \tilde{K}(z, A)$ is measurable as a composition of measurable functions.
This makes $\tilde{K}$ a valid Markov kernel.
Suppose $\pi$ is an invariant measure for $K$ on $\mathcal{X}$, i.e., 
\begin{align}
    \pi(B) = \int_{\mathcal{X}} K(x, B)\pi(dx),\quad \forall B \in \mathcal{F}.
\end{align}
Define the pushforward measure $\rho \triangleq \pi \circ f^{-1}$ on $(\mathcal{Z}, \mathcal{G})$.
Using $x = f^{-1}(z)$, we have for any measurable $A \in \mathcal{G}$:
\begin{align}
    \rho(A) = \pi(f^{-1}(A)) = \int_{\mathcal{X}} K(x, f^{-1}(A)) \pi(dx)  = \int_{\mathcal{Z}}\tilde{K}(z,A)\rho(dz),
\end{align}
so $\rho$ is invariant under $\tilde{K}$.

In practical MCMC, this means that we can sample from a transformed target distribution instead of the original target. Having obtained samples $z_1, \dots, z_n$ from the transformed target in space $\mathcal{Z}$, we simply transform them to space $\mathcal{X}$ via $x_i = f^{-1}(z_i)$ for $i = 1, \dots, n$.
Such sampling is termed preconditioned MCMC.
In typical sampling contexts where $(\mathcal{X}, \mathcal{F}), (\mathcal{Z}, \mathcal{G})$ are standard Borel spaces, any bijective measurable map $f$ has a measurable inverse $f^{-1}$~\cite[Corollary 15.2, page 89]{kechris_classical_1995}.
This includes classical affine map preconditioners of the form $f(x) = L^{-1}(x-\mu)$ where $x, \mu \in \mathbb{R}^d, L \in \mathbb{R}^{d \times d}$, $d$ is the target dimensionality, and $L$ is lower-triangular (see~\cite{hird_quantifying_2025} for a recent analysis).

The same mathematical framework also allows for nonlinear maps $f$, including invertible neural networks as components of normalizing flows~\citep{hoffman_neutralizing_2019}.
Normalizing flows (NF) are distributions, defined as transformations of a simple distribution via an invertible deep neural network.
The log density $\log q(x)$ of an NF $Q$ can be explicitly computed with the change-of-variables formula:
\begin{align}
    \log q(x) = \log p_Z(f(x)) + \log |\det J_f (x)|,
\end{align}
where $p_Z$ is the density of the latent distribution $Z$ (typically standard Gaussian), $f:\mathbb{R}^D \rightarrow \mathbb{R}^D$ is the forward map of the invertible neural network, and $J_f$ is the Jacobian matrix of $f$, evaluated at $x$.
This explicit density formula allows for straightforward optimization in the contexts of stochastic variational inference and maximum likelihood estimation, which is key to performing model updates in preconditioned MCMC.

\subsection{Approximate Gaussianity heuristic}
Let $\pi$ be a continuous distribution in $\mathbb{R}^D$ and let $x_1, \dots, x_n$ be independent and identically distributed (IID) samples from $\pi$.
For a dimension $k \in [D]$, we wish to determine whether the $k$-th marginal distribution of $\pi$ is approximately Gaussian.
We approach this using the 2-Wasserstein distance.

Let $x_1^{(k)}, \dots, x_n^{(k)}$ be the $k$-th dimension of samples from $\pi$, giving rise to an empirical distribution $\tilde{\pi}^{(k)}$.
Let $\tilde{Q}^{(k)}: [0, 1] \rightarrow \mathbb{R}$ be the quantile function of $\tilde{\pi}^{(k)}$ and let $\Phi^{-1}: [0, 1] \rightarrow \mathbb{R}$ be the analytically-known standard Gaussian quantile function.
The 2-Wasserstein distance between $\tilde{\pi}^{(k)}$ and the standard normal is:
\begin{align}
    W_2(\tilde{\pi}^{(k)}, N(0, 1)) = \left(\int_0^1 |\tilde{Q}^{(k)}(\psi) - \Phi^{-1}(\psi)|^2 d\psi\right)^{1/2}.\label{eqn:wd-to-gauss-exact}
\end{align}
Since $\tilde{\pi}^{(k)}$ is an empirical distribution, its quantile function $\tilde{Q}^{(k)}$ is a piecewise constant function that takes values from sorted samples $x_1^{(k)}, \dots, x_n^{(k)}$.
The sorted samples are order statistics $x^{(k)}_{(1)}, \dots, x^{(k)}_{(n)}$, where the $i$-th order statistic corresponds to the empirical quantile near $\psi = i/n$ for $i = 1, \dots, n$. In our implementation, we instead choose the interpolation $\psi = (i - 0.5) / n$ that centers the quantile level in the middle of each empirical jump~\citep{hyndman_sample_1996}.
We can now approximate the integral as:
\begin{align}
    W_2(\tilde{\pi}^{(k)}, N(0, 1)) \approx \left(\frac{1}{n} \sum_{i=1}^n \left|x^{(k)}_{(i)} - \Phi^{-1}\left(\frac{i-0.5}{n}\right)\right|^2\right)^{1/2}.
\end{align}
$\tilde{Q}^{(k)}$ converges almost surely to the true $k$-th marginal quantile function of $\pi$ by the Glivenko-Cantelli theorem, ensuring the approximation becomes close for large $n$.
We also standardize samples before computing the distance, which makes $W_2$ the same regardless of the location and scale of $\tilde{\pi}^{(k)}$, instead only focusing on the shape of the distribution.

We consider a dimension approximately Gaussian if the $W_2$ distance does not exceed a threshold $\tau$.
We note that the error in estimating $W_2$ in terms of sample size $n$ converges to zero with $\mathcal{O}(n^{-1/2})$~\citep{fournier_rate_2015}.
We thus set $\tau = C+K/\sqrt{n}$, where $C$ determines the threshold for approximate Gaussianity and $K$ handles the uncertainty in estimating $W_2$ with small $n$.
According to~\cite{irpino_basic_2015}, the \textit{squared} 2-Wasserstein distance can be decomposed into terms involving the location, scale, and shape of two input distributions $p$ and $q$:
\begin{align}
    W_2(p, q)^2 = (\mu_p - \mu_q)^2 + (\sigma_p - \sigma_q)^2 + 2\sigma_p \sigma_q(1 - \rho_{p, q}),\label{eqn:wd-decomposition}
\end{align}
where $\mu_p, \mu_q \in \mathbb{R}$ are the means of $p$ and $q$, $\sigma_p, \sigma_q>0$ are their standard deviations, and $\rho_{p, q} \in [0, 1]$ is the Pearson correlation of the points in the quantile-quantile plot of $p$ and $q$.
We note that the correlation is non-negative as the quantile function is non-decreasing.
Since the means and standard deviations are equal in our case, both standard deviations are $1$, and the correlation is between 0 and 1, the distance can be simplified to $W_2(p, q)^2 = 2 (1 - \rho_{p, q})$, which implies $W_2(p, q)^2 \leq 2$ and therefore $W_2(\tilde{\pi}^{(k)}, N(0, 1)) \leq \sqrt{2}$.
This further implies that the underlying standard deviation of $W_2(\tilde{\pi}^{(k)}, N(0, 1))$ is at most $\sqrt{2} / 2$.
By setting $K = \Phi^{-1}(\alpha)\sqrt{2}/2$, we construct at least a confidence interval of size at least $\alpha$.
We set $\alpha = \Phi(2) \approx 0.977$ in our experiments, which implies $D = \sqrt{2}$.
The resulting threshold is thus $\tau = C+\sqrt{2/n}$, which results in a probability of at most 2.3\% that $\tilde{\pi}^{(k)}$ is approximately Gaussian, but is falsely identified as not being such.
We investigate sensitivity to $C$ using synthetic distribution testing (see Appendix~\ref{app:sec:threshold} for a discussion).

\subsection{Factorized normalizing flow preconditioner}
We define the factorized NF preconditioner with a joint distribution dealing with two disjoint subsets of variables.
Let $[D]$ denote the set of integers $\{1, \dots, D\}$.
Let $x \in \mathbb{R}^D$ be a point in the support of the target distribution $\pi$.
Let $G = \{j_1, \dots, j_m\}, j_k \in [D]$ be the set of indices corresponding to Gaussian dimensions and $H = [D] \setminus G = \{\ell_1, \dots, \ell_{D-m}\}$ be the set of indices corresponding to non-Gaussian dimensions.
The proposed joint density $p(x)$ for a point $x$ is:
\begin{align}
    x_G &= (x^{(j_1)}, \dots, x^{(j_m)}), \\
    x_H &= (x^{(\ell_1)}, \dots, x^{(\ell_{D-m})}), \\
    p(x) &= p_G(x_G)p_H(x_H | x_G),
\end{align}
where $p_G$ is models the approximately Gaussian dimensions, and $p_H$ the non-Gaussian dimensions.

We model the approximately Gaussian dimensions with $x_G \sim N(\mu, \Sigma)$ where $\mu \in \mathbb{R}^m$ is the empirical mean vector of $m$ Gaussian dimensions and $\Sigma \in \mathbb{R}^{m \times m}$ is the corresponding empirical covariance matrix.
Finding the optimal preconditioner for approximately Gaussian data can be prone to overfitting on small datasets.
Furthermore, in the time it takes to fit such a preconditioner, we could simply obtain better training data, which would both facilitate training the preconditioner for non-approximately Gaussian dimensions and help better identify approximately Gaussian dimensions.
Instead, the choice of a multivariate normal distribution allows for relatively accurate, stable, and fast fitting to data.
The corresponding linear map $f_G^{-1}$ that transforms latent points $z_G$ to target points $x_G$ is defined as:
\begin{align}
    x_G = f_G^{-1}(z_G) = Lz_G + \mu,
\end{align}
where $LL^\top = \Sigma$ is the Cholesky decomposition of the previously noted empirical covariance matrix of $x_G$ and $\mu$ is the same as described above.

We model the remaining dimensions with a conditional NF, which consists of a standard Gaussian latent distribution and a map $f(\cdot | y) : \mathbb{R}^{D-m} \rightarrow \mathbb{R}^{D-m}$ that is bijective in the first argument.
By conditioning the map on approximately Gaussian dimensions, we reduce the difficulty of training the NF compared to trying to fit all dimensions.
Our formulation is general and allows for various conditional NF architectures.
We abbreviate the factorized NF preconditioner as FF.

In our experiments, we consider the real non-volume-preserving flow~\citep[RNVP;][]{dinh_density_2017}.
We define the RNVP forward map by a composition of bijective layers:
\begin{align}
    f = f_{\mathrm{ActNorm}}^{(0)} \circ b_1 \circ \dots \circ b_k,
\end{align}
where $f_\mathrm{ActNorm}^{(0)}$ is a trainable elementwise affine map which is initialized to a standardization map using the first training batch, i.e., $x \mapsto (x - \mu_b) / \sigma_b$ where $\mu_b$ and $\sigma_b$ are the empirical mean and standard deviation vectors of the first batch.
Each block $b_i, i=1,\dots, k$ is a composition of three bijections:
\begin{align}
    b_i = f_{\mathrm{perm}}^{(i)} \circ f_{\mathrm{coupling}}^{(i)} \circ f_{\mathrm{ActNorm}}^{(i)},
\end{align}
where $f_\mathrm{perm}^{(i)}$ is a permutation matrix, and $f_{\mathrm{coupling}}^{(i)}$ is an affine coupling bijection.
Each permutation matrix reverses the order of dimensions.

The forward map for the unconditional affine coupling bijection is defined by first partitioning an input $x_H$ into $x_H^A \in \mathbb{R}^{|A|}$ and $x_H^B \in \mathbb{R}^{|B|}$ where $A$ and $B$ are disjoint sets of $D - m$ indices for non-approximately Gaussian dimensions.
We partition an input by splitting it in half across dimensions.
The transformation proceeds as $y_H^A = x_H^A$ and $y_H^B = \alpha x_H^B + \beta$, where $\alpha, \beta$ are the elementwise affine map parameter vectors, typically predicted by a conditioner neural network from input $x_H^A$.
We condition this bijection on approximately Gaussian dimensions by passing them to the conditioner neural network as auxiliary inputs.
In this paper, we use a single linear map as the conditioner that takes as input the concatenation of $x_H^A$ and $x_G$ and predicts two parameter vectors with the same shape as $x_H^B$:
\begin{align}
    \log \alpha, \beta = \mathtt{unstack}(W^\top \mathtt{stack}(x_H^A, x_G) + b),
\end{align}
where the \texttt{stack} operation appends the rows of $x_G$ to $x_H^A$, \texttt{unstack} splits the output down the middle into two equally-sized vectors, $W \in \mathbb{R}^{(|A| + m)\times 2 |B|}$ is the weight matrix, and $b \in \mathbb{R}^{2|B|}$ is the bias vector.
Using a linear preconditioner means fewer trainable parameters, making the model less prone to overfitting.
Each block $b_i$ remains expressive through the implicit nonlinear exponential transformation of the predicted parameter $\log \alpha$.

\section{Experiments}
In this section, we demonstrate how a factorized RNVP (F-RNVP) architecture improves MCMC sample quality compared to a classic RNVP architecture, as well as diagonal and triangular linear preconditioners.
We first demonstrate improved sampling performance for synthetic targets where a subset of dimensions is exactly Gaussian by design.
We then explore performance for real-world Bayesian model posteriors without exact Gaussianity.
First, we analyze how posterior samples change as we vary the strengths of the likelihood and prior.
We then analyze HBMs with very weak likelihoods, which exhibit strong funnel geometries.
Our NF training and step size tuning procedure follows a cycle-based approach, commonly used in popular MCMC software~\citep{carpenter_stan_2017}.
We define it precisely in Appendix~\ref{app:sec:warmup}, along with MCMC and NF hyperparameter choices (unless stated otherwise for different experiments).

Both RNVP and its factorized counterpart use the initial ActNorm layer and two previously defined blocks. Both also use the linear conditioner in affine coupling layers.
Each HMC kernel uses an initial step size of 0.01, similar to the parameterization by~\cite{nabergoj_empirical_2024}.
\cite{hoffman_neutralizing_2019} use 4 leapfrog steps to show stark differences between NeuTra HMC and linear preconditioning. 
We opt for 20 leapfrog steps to allow for better MCMC samples with linear preconditioning, as this evens the balance between slow but expressive NF preconditioning and fast but inexpressive linear preconditioning.
The initial state distribution is set to an independent $\mathrm{Uniform}(-2, 2)$ for each dimension, as used in the robust sampling package Stan~\citep{carpenter_stan_2017}.
Kernels always operate in an unconstrained space $\mathbb{R}^D$ for a $D$-dimensional target distribution.
We compute positive model parameters by transforming the corresponding unconstrained parameter with the softplus function.
For parameters on an interval, we achieve this with the sigmoid function, followed by a scale and shift.

\subsection{Sampling tails of synthetic distributions}
We analyze two synthetic distributions that represent different approximate Gaussianity scenarios.
These are the Neal's funnel distribution and a banana distribution.
The first is a 10-dimensional distribution with a single Gaussian marginal, defined as:
\begin{align}
    x_0 &\sim N(0, 3^2), \\
    x_i | x_0 &\sim N(0, \exp(x_0 / 2)^2)\; \mathrm{for} \; i = 1,\dots,9. \label{eqn:neals-funnel}
\end{align}
The second is a 100-dimensional distribution with 99 Gaussian marginals:
\begin{align}
    x_0 &\sim N(0, 10^2), \\
    x_1 | x_0 &\sim N((3/100) x_0^2 - 3, 1), x_i \sim N(0, 1)\;\mathrm{for} \; i = 2,\dots,99. \label{eqn:banana}
\end{align}

\begin{figure}[htb]
    \centering
    \includegraphics[width=0.8\linewidth]{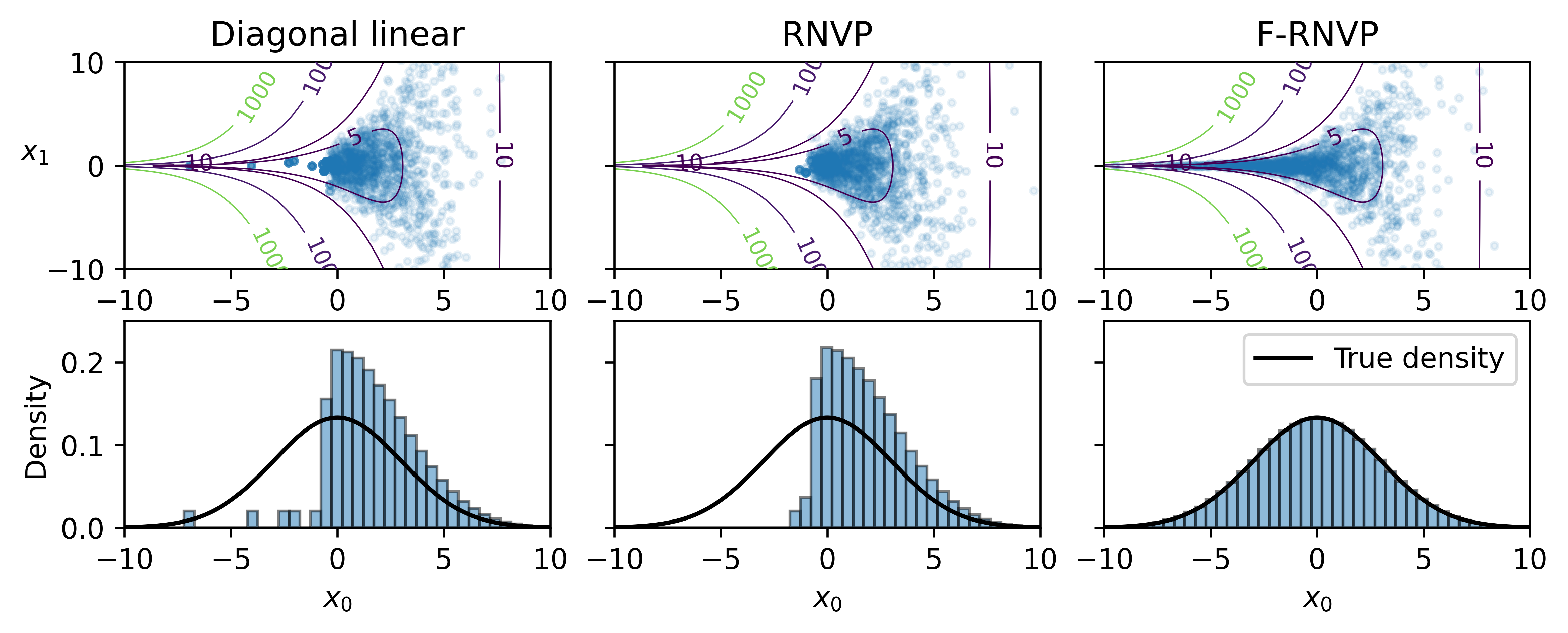}
    \caption{HMC samples from Neal's funnel distribution with different preconditioners: a diagonal linear map, RNVP, and F-RNVP. The scatterplots show samples for random variables $x_0$ and $x_1$. Contour lines show the corresponding true negative log probability density value. The histograms represent the right tails of empirical densities for $x_0$. Note that the contours for 5, 10, and 100 are truncated on the left due to numerics, but actually extend towards $-\infty$.}
    \label{fig:funnel}
\end{figure}

Neal's funnel is a challenging distribution to sample as most of the probability mass is concentrated in a thin tail with low volume, while a region with much greater volume has comparatively very small mass.
It has previously been observed that a single step size is inadequate to deal with this volume change when traversing between the two regions via local MH (e.g., HMC) transitions.
The NF preconditioning approach termed NeuTra HMC~\citep{hoffman_neutralizing_2019} transforms the distribution into a Gaussian, which aims to remove the pathology.
In Figure~\ref{fig:funnel}, we compare HMC draws when using a diagonal linear preconditioner, as well as the RNVP and our proposed factorized counterpart.
Note that our HMC preconditioning method uses a cycle-based warmup strategy, which differs from the SVI strategy in the original NeuTra HMC formulation.

While the scatterplot provides a coarse view suggesting similar draws across the three cases, the marginal distributions reveal stark differences.
F-RNVP allows HMC to successfully explore the thin end of the funnel, whereas the other two preconditioners cannot facilitate exploration for $x_0 < 0$.
Visually, its empirical marginal density for $x_0$ matches the ground truth nearly perfectly.
The few stranded samples corresponding to the diagonal map and RNVP at the thin end indicate that the chains are stuck.
Compared to F-RNVP, their preconditioned spaces do not balance the two volumes or allow transitions with a single HMC step size.
The large discrepancy can be attributed to the designed parameterization of our model.
While independent modeling via the diagonal map and fully joint modeling via RNVP result in under- and overparameterized preconditioners, respectively, our approach correctly identifies the first dimension as approximately Gaussian.
This creates a conditional NF, whose dependence structure is much better suited to the training data.

\begin{figure}[htb]
    \centering
    \includegraphics[width=0.8\linewidth]{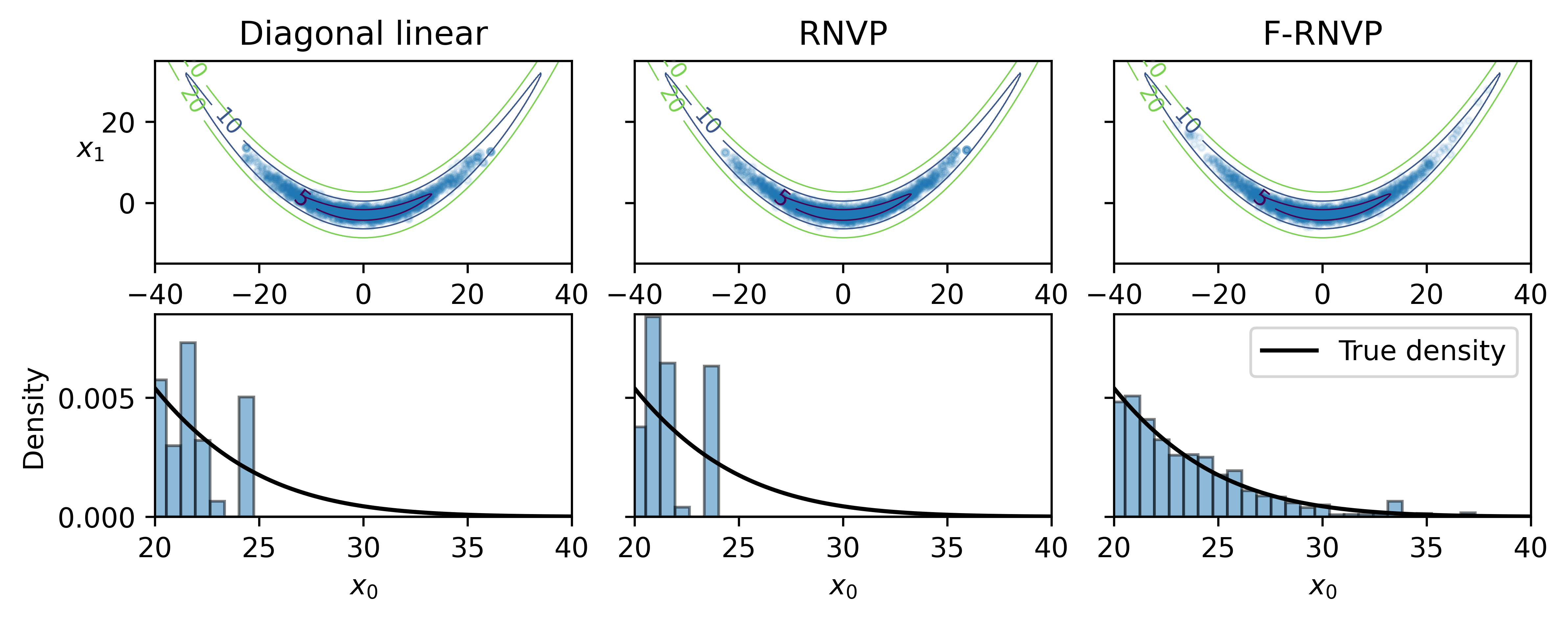}
    \caption{HMC samples from the banana distribution with different preconditioners: a diagonal linear map, RNVP, and F-RNVP. The scatterplots show samples for random variables $x_0$ and $x_1$. Contour lines show the corresponding true negative log probability density value. The histograms represent the right tails of empirical densities for $x_0$.}
    \label{fig:banana}
\end{figure}

We repeat the analysis for the banana target and visualize results in Figure~\ref{fig:banana}.
Our proposed model successfully samples from the tails of the distribution, whereas the competing approaches are stuck in the interval of roughly $(-20, 20)$.
The tails account for approximately 4.5\% of total marginal probability mass and represent more extreme distribution samples.
The histograms show that F-RNVP models the tails more efficiently than the competing approaches.
The heuristic correctly identifies 99 dimensions as approximately Gaussian.
This substantially simplifies the conditional NF to only model one dimension via an affine map whose parameters are informed by the identified approximately Gaussian dimensions.
Referring to Equation~\ref{eqn:banana}, the NF only needs to assign zero weight to contributions from $x_2, \dots, x_{99}$ while these remain constant.
In contrast, full RNVP preconditioning involves simultaneous nonlinear learning of the otherwise standard Gaussians and inference of the dependence structure, which hinders the training process.

\subsection{Sparse logistic regression posterior analysis}\label{subsec:sgc}
We evaluated our approach on HMC sampling from the sparse German credit posterior, also analyzed in the original NF preconditioning paper~\citep{hoffman_neutralizing_2019}.
This is a sparse logistic regression model with $m = 51$ parameters, whose likelihood is:
\begin{align}
    y_i &\sim \mathrm{Bernoulli}(\sigma(\tau (\boldsymbol{\beta} \odot \boldsymbol{\lambda})^\top \mathbf{x}_i)),
\end{align}
where $\sigma$ is the sigmoid function. The model parameters are $\tau \in \mathbb{R}_+, \boldsymbol{\lambda} \in \mathbb{R}^m_{+}, \boldsymbol{\beta} \in \mathbb{R}^m_+$.
The priors are $\boldsymbol{\beta} \sim N(0, I)$ and $\tau, \lambda^{(j)} \sim \mathrm{Gamma}(0.5, 0.5)$ for $j=1,\dots, m$ using the shape-rate Gamma parameterization. The $\boldsymbol{\lambda}$ parameters introduce sparsity to the model.
We sample $\tau^\prime \in \mathbb{R}$ and $\boldsymbol{\lambda}\prime \in \mathbb{R}^m$ in the unconstrained real space and use $\tau = \log (1 + \exp(\tau^\prime)), \lambda^{(j)} = \log (1 + \exp(\lambda^{\prime(j)}))$ to compute the posterior density for HMC sampling. 
We considered the diagonal and dense (i.e., triangular) linear preconditioners, RNVP, and its factorized counterpart.
We included the dense linear preconditioner, as observing posterior samples on preliminary runs showed moderate-to-high correlations for several parameter pairs.

\subsubsection{Impact of prior vs likelihood strength}

The original dataset consists of $n = 1000$ data points.
For F-RNVP to be applicable in small-data and large-data regimes, it should perform as well or better than other preconditioners regardless of $n$.
We experiment with reducing this number, thereby increasing the contribution of the prior.
As 25 of 51 parameters have a standard Gaussian prior, we expect our approach to identify more approximately Gaussian dimensions for smaller $n$.
While the prior does not explicitly encode a funnel structure, the likelihood contributes substantially to a geometrically complex posterior.
Regardless of whether the posterior is dominated by the likelihood or the prior, we expect that identifying approximately Gaussian dimensions leads to better preconditioner training and thus also sampling.

\begin{figure}[htb]
    \centering
    \includegraphics[width=0.9\linewidth]{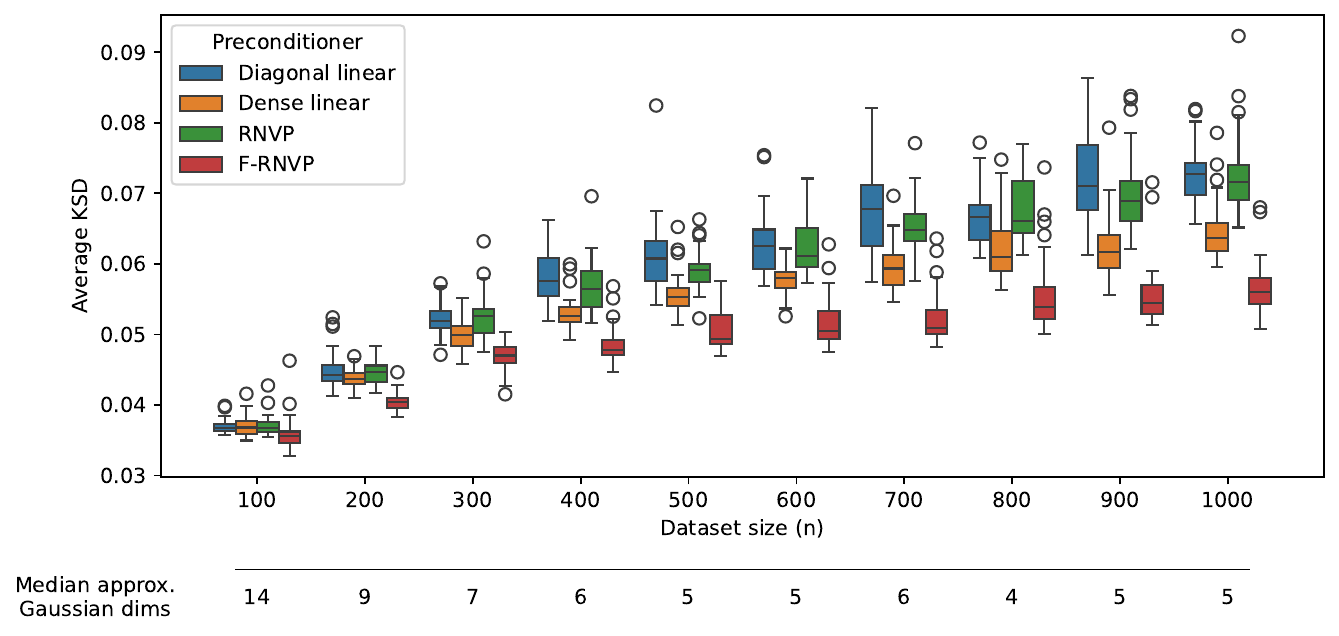}
    \caption{Distributions of average KSD for different preconditioners and dataset sizes $n$ on the sparse German credit dataset. Boxplots are based on 30 experiments. Colored boxes denote the inter-quartile range. The average KSD in each experiment is computed with 250 random HMC samples over 150 trials. This measure incorporates many MCMC samples, avoiding the prohibitive $O(n^2)$ time and space complexity of computing KSD. The numbers below dataset size ticks represent the median number of identified approximately Gaussian dimensions according to the F-RNVP preconditoner. KSD values are not directly comparable across different dataset sizes, as each size affects the likelihood and thus the posterior. This visualization compares preconditioners across different $n$ while preserving absolute values useful for future comparisons.}
    \label{fig:sgc-boxplot}
\end{figure}

We ran experiments with different random seeds for each dataset size $n \in \{100, 200, \dots, 1000\}$, resulting in randomized initial chain states and RNVP initialization.
We computed KSD given HMC draws and target score functions.
We show the boxplot of these values across 30 differently-seeded experiments in Figure~\ref{fig:sgc-boxplot}.

We find that the boxplot median is always lower for F-RNVP than for other preconditioners, indicating that our approach has a better median performance regardless of likelihood/prior prevalence.
For $n \geq 200$, its inter-quartile range (IQR) does not overlap with the IQR of any other preconditioner, suggesting that our approach is highly probable to achieve a lower KSD than other preconditioners.

\subsubsection{Visually investigating tail behavior}
By observing two-dimensional HMC draw projections, the posterior with $n = 1000$ exhibits an approximately Gaussian distribution with some correlated dimensions.
This explains the better performance of the dense linear preconditioner relative to the diagonal preconditioner in Figure~\ref{fig:sgc-boxplot}.
The classic RNVP model often performs similarly to the diagonal linear preconditioner.
We observed the learned RNVP distribution on some HMC runs and found it largely similar to a diagonal Gaussian, a consequence of overparameterization.
Importantly, we found that our model identifies several dimensions as not approximately Gaussian, which leads to a more suitable NF parameterization, while retaining the benefits of dense linear preconditioning on the remaining dimensions.

\subsection{Funnel sampling in multi-level posteriors with weak likelihoods}
We consider four HBM targets where the prior explicitly encodes a funnel shape.
The posteriors retain this shape when the likelihood is weak, i.e., there are few data points.
This small-data phenomenon was highlighted by~\cite{betancourt_hamiltonian_2015} as posing substantial difficulties for MCMC samplers due to the inherent funnel structure.

\subsubsection{Experiment setup}
We modify the sparse German credit model (see Section~\ref{subsec:sgc}) and three radon models below to use Gaussian priors with a log-variance parameter.
For sparse German credit, this means adding $\sigma \sim \mathrm{LogNormal}(0, 1)$ and replacing the $\beta$ prior with $\beta \sim N(0, \sigma I)$.
This modified model explicitly relates to Equation~\ref{eqn:neals-funnel}, where $x_0$ can be interpreted as the log standard deviation parameter and $x_i$ corresponds to logistic regression coefficients $\beta_i$.

The radon targets are hierarchical Bayesian regression models that describe the quantity of radon in Minnesotan households based on the house index $i$, the county $c(i) \in [85]$, and the floor $f(i) \in \{0, 1\}$ where the measurement was taken.
The likelihood equals $y_i \sim N(\mu_i, \sigma_y)$, where the Gaussian mean is computed differently in each model:
\begin{align}
    \mu_i &= a_{c(i)} f(i) + b \tag{\textit{varying slopes}; R-VS}, \\
    \mu_i &= a f(i) + b_{c(i)}  \tag {\textit{varying intercepts}; R-VI}, \\
    \mu_i &= a_{c(i)} f(i) + b_{c(i)} \tag {\textit{varying slopes and intercepts}; R-VSI}.
\end{align}
In all models, the data standard deviation has the prior $\sigma_y \sim \mathrm{LogNormal}(0, 1)$.
In the R-VS model, the slope priors are $\mu_a \sim N(0, (10^5)^2)$, $\sigma_a \sim \mathrm{LogNormal}(0, 1)$, $a_{c(i)} \sim_{\mathrm{IID}} N(\mu_a, \sigma_a^2)$ and the intercept prior is $b \sim N(0, (10^5)^2)$
In the R-VI model, the intecept priors are $\mu_b \sim N(0, (10^5)^2)$, $\sigma_b \sim \mathrm{LogNormal}(0, 1)$, $b_{c(i)} \sim_{\mathrm{IID}} N(\mu_b, \sigma_b^2)$, and the slope prior is $a \sim N(0, (10^5)^2)$.
In the R-VSI model, the slope priors are taken from R-VS, and the intercept priors are taken from R-VI.
Parameter combinations $(\sigma_a, a_{c(i)})$ and $(\sigma_b, b_{c(i)})$ for $i \in [85]$ create a funnel shape similar to Equation~\ref{eqn:neals-funnel}.

To demonstrate that our method works on weak likelihood posteriors, we used the first row in the original two datasets as the only data point in the likelihood.
Given this setup, a successful sampling run should yield samples resembling a slightly modified prior with an approximate funnel shape.
We ran the preconditioned no-U-turn sampler~\citep[NUTS;][]{hoffman_nouturn_2011} for 1000 sampling iterations, similar to typical applied Bayesian analyses.
NUTS simulates Hamiltonian dynamics similar to HMC, but automatically adapts its trajectory length to local posterior geometry.
We used 100 parallel chains as in~\citep{nabergoj_empirical_2024}, which yielded varied draws to train NF preconditioners and facilitate covariance estimation for linear preconditioners.
We used $C = 0.1$ as the approximate Gaussianity constant for the sparse German credit posterior and $C = 0.01$ for the radon posteriors ($C = 0.1$ gave similar, but slightly worse results for radon).
We compared RNVP, F-RNVP, and the diagonal preconditioner.

While reasonable for well-posed distributions, we noticed that KSD is often problematic for these strongly funnel-shaped posteriors.
This is due to its reliance on pairwise distances between draws, which vary greatly between the thin and thick ends of the funnel.
We found that KSD with the Gaussian RBF kernel or the inverse multiquadric kernel often assesses draws as high quality even if scatterplot and traceplot heuristics indicate substantial mixing problems (relative to visually well-mixed chains).
For this reason, we instead evaluate preconditioners using effective sample size (ESS).
We observe the minimum ESS, which corresponds to worst-case performance, i.e., relating to the most challenging dimensions.
We use bulk ESS and tail ESS (using 5\% and 95\% quantiles) as defined by~\cite{vehtari_ranknormalization_2021}.
Assuming good exploration, bulk ESS describes sampler behavior for a greater region of the posterior.
Meanwhile, tail ESS describes behavior in the thin end of the funnel, i.e., the most challenging sampling region.

\subsubsection{Results}\label{subsubsec:results-onept}
\begin{figure}[htb]
    \centering
    \includegraphics[width=0.9\linewidth]{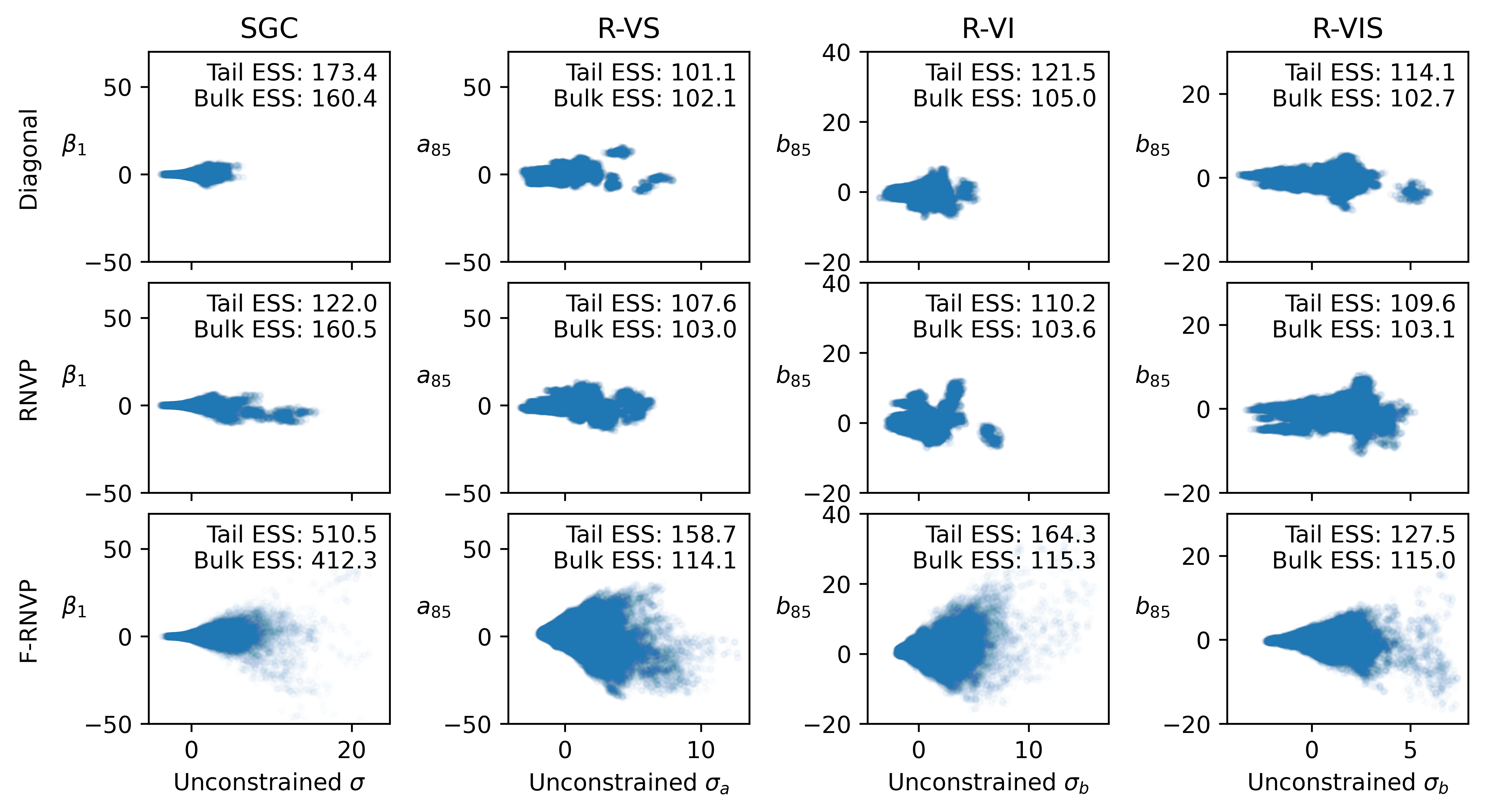}
    \caption{ESS and scatterplots of NUTS draws for different combinations of posteriors in columns and preconditioners in rows. All posteriors use a single data point for the likelihood. Scatterplots show draws for the unconstrained standard deviation and a parameter that uses it in its Gaussian prior. The shown ESS values represent minima across all dimensions and based on 1000 sampling iterations.}
    \label{fig:one-point}
\end{figure}

We visualize posterior samples in Figure~\ref{fig:one-point}.
F-RNVP always achieves higher ESS than the RNVP and diagonal preconditioners.
Importantly, its tail ESS is much higher due to better mixing in the thin end of the funnel.
The bulk ESS is also higher, though the differences are smaller.
Preconditioning with F-RNVP leads to scatterplots with an apparent funnel shape.
This is crucial for preconditioner evaluation, as using a single data point with relatively narrow $\mathrm{LogNormal}(0, 1)$ priors for standard deviations preserves the funnel shape in these posteriors.
The performance improvement in tail modeling here is consistent with our synthetic experiments in Figure~\ref{fig:funnel} and Figure~\ref{fig:banana}.

\begin{figure}[htb]
    \centering
    \includegraphics[width=0.9\linewidth]{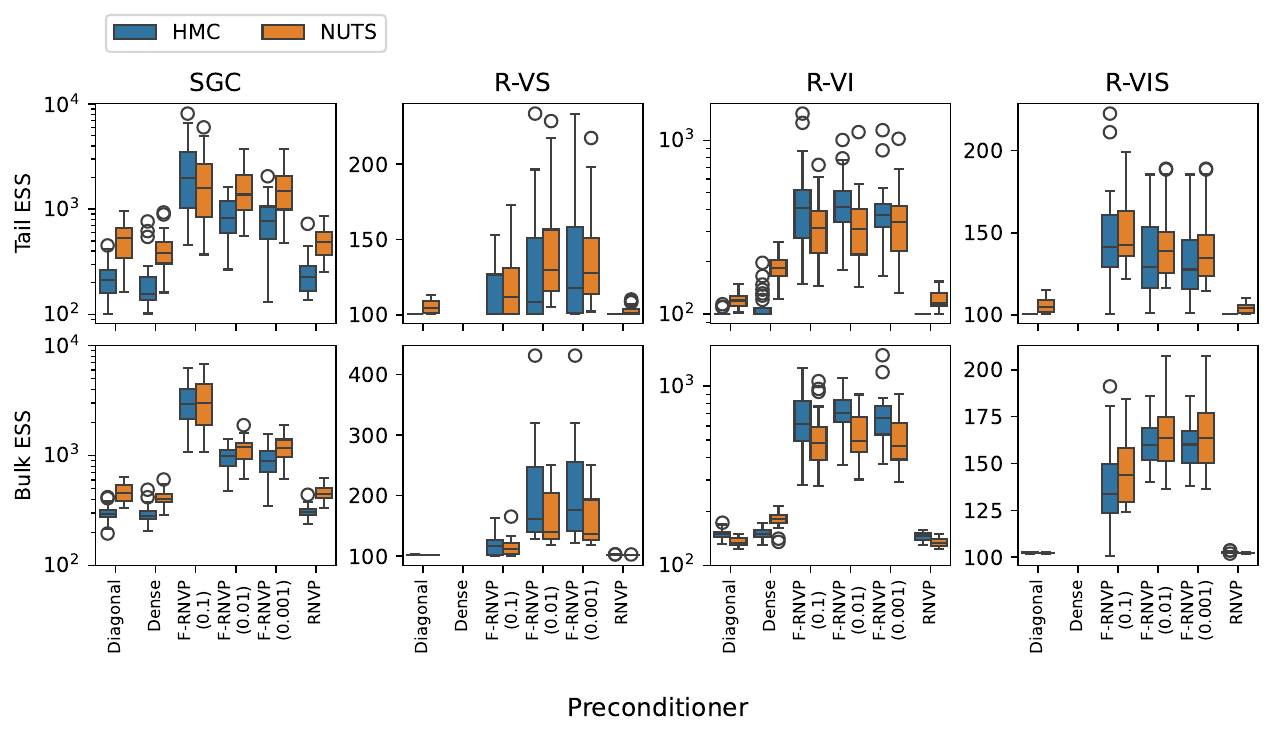}
    \caption{Tail ESS and bulk ESS for different combinations of target distributions and preconditioners. All likelihoods utilize a single dataset row. Boxplots are based on 30 repeated runs with different random seeds. SGC and R-VI columns use log scales. The shown ESS values represent minima across all dimensions.}
    \label{fig:ess}
\end{figure}

We repeated the experiments with 30 different random seeds for each combination of preconditioner and target distribution to account for uncertainty and provide a numeric comparison.
We also performed the same experiments for HMC to compare performance.
In Figure~\ref{fig:ess}, we visualize tail ESS and bulk ESS, computed with states from all 100 chains and 1000 steps.

Fitting the dense linear preconditioner has stability issues on R-VS and R-VIS targets. Their high dimensionality and complex geometry hinder the stability of computing the corresponding lower-triangular transformation matrix. We nevertheless include the dense preconditioner in the plots for completeness.

The median ESS is always higher with F-RNVP preconditioners than with other methods.
While F-RNVP sometimes attains ESS with high uncertainty, ESS is favorably concentrated at values above the median in several of these cases.
The choice of the approximate Gaussianity constant $C$ affects sampler results, e.g., when observing bulk ESS for the SGC and R-VS targets.
A high value of $C$ is suitable for some targets, but not for others.
The selection is linked to the targeted posterior distribution.
We note that the preconditioners are ordered from left to right according to increasing model complexity.
F-RNVP is a suitable middle ground between the simple diagonal preconditioner and the complex RNVP preconditioner, both of which are special cases of our model.
NUTS attains a higher ESS than HMC with diagonal preconditioning, which is reasonable due to its adaptive trajectory length selection.
However, the differences between the two samplers appear much smaller when using the F-RNVP preconditioner.
This is reasonable because the preconditioner transforms the target distribution to an approximately uncorrelated Gaussian.
In such a space, the default number of HMC trajectory steps enables sufficient exploration.
The adaptive trajectory length selection in NUTS does not substantially change how the target is explored.

\section{Conclusion}
We presented an approach for reducing NF preconditioner complexity in MCMC by independently modeling approximately Gaussian dimensions of a target distribution.
Using a Wasserstein distance–based heuristic, we identified these dimensions and showed that the associated threshold estimator converges to the true threshold at a rate of $\mathcal{O}(n^{-1/2})$, where $n$ is the number of MCMC draws.

We implemented our method with the RNVP architecture, factorized into an approximately Gaussian component and a conditional NF.
We evaluated F-RNVP on benchmark targets, including the banana distribution and Neal’s funnel, as well as on a sparse logistic regression model. 
In all cases, it produced higher-quality samples, particularly in the tails, than both the non-factorized RNVP and the diagonal linear preconditioner.

Furthermore, we demonstrated that the proposed approach improves sampling quality for HBMs with strong funnel geometries, explicitly enforced by using only a single likelihood data point and a standard deviation parameter to relate to Neal's funnel formulation.
For targets with strong funnel geometries, F-RNVP achieved higher ESS and produced posterior samples that better reflected the true geometry, demonstrating enhanced exploration efficiency. 
These gains remained across choices of the Gaussianity constant, indicating that the method is not sensitive to hyperparameter tuning.

The performance gains stem from explicitly separating approximately Gaussian and non-Gaussian dimensions, which simplifies preconditioner training and improves stability. 
These results position F-RNVP as an effective and general-purpose preconditioner for HBM analyses, capable of adapting from complex posteriors to approximately Gaussian regimes where it naturally reduces to linear preconditioning.
F-RNVP can also reduce the need for more complex MCMC methods, i.e., the performance of NUTS and HMC was not markedly different.

\subsection{Limitations}
The main consideration in our approach is the selection of the approximate Gaussianity constant $C$.
If a large number of MCMC draws are available to train the preconditioner, and the preconditioner training is computationally fast relative to MCMC exploration, then it would be practical to reduce the constant.
Currently, our recommendation for such scenarios is to repeat MCMC runs with a geometric sequence of constants, e.g., $C = 10^{-k}$ for $k = 1, \dots, k_\mathrm{max}$, which can be time-consuming.
Our recommended options, $C = \{0.1, 0.01, 0.001\}$, have proven useful in our experiments and indeed improve performance over non-factorized preconditioners, but they are not general.
Future work could consider a scheme that determines the approximate Gaussianity constant from the warm-up procedure, rather than treating it as a hyperparameter.

Another limitation is that the factorized NF preconditioner is rebuilt at the start of each warmup cycle.
This means discarding potential progress on parameter optimization, which could be important when dealing with highly local chains.
In such a scenario, a chain would explore one posterior region in one cycle and a different one in the next.
We mitigate the information loss by using a reservoir of training samples, which preserves samples from the entire warmup process.

We evaluated our approach using the RNVP architecture, as it is commonly used in MCMC with NFs.
The key to improving results lies in factorizing the preconditioner and delegating nonlinear modeling to the conditional NF.
The approximate Gaussianity heuristic is the primary component governing the complexity of the preconditioner. 
This is especially noticeable when the number of identified approximately Gaussian dimensions is small, which leads to a smaller conditional NF.
As smaller models are typically easier to fit, the choice of the architecture is not crucial.
In cases with few approximately Gaussian dimensions, we can consider two cases.
In broad terms, if the remainder of the distribution is not very complex, then the choice of conditional NF architecture has relatively little impact on overall sampling efficiency.
However, the architecture choice could have a greater impact if the remainder of the distribution has a highly complex geometry.
This is not generally the case for funnel geometries in HBMs, the focus of experiments in this work, where the nonlinear dimensions can often be approximately Gaussian conditional on the unconstrained standard deviation variable.

Our framework can also be extended to consider other distributions via the Wasserstein distance heuristic, such as the Student's t or the lognormal distribution.
We would retain the Wasserstein distance heuristic and replace the Gaussian quantile in Equation~\ref{eqn:wd-to-gauss-exact} with that of Student's t or the lognormal distribution.
Assuming the reference univariate distribution has a finite mean and variance, the decomposition in Equation~\ref{eqn:wd-decomposition} will remain and result in a controlled range of constants $C$.
If considering several candidate distributions with finite moments, we would choose the one with the highest probability density of marginal empirical draws.
We delegate this to future work.

\acks{This work was supported by the Slovenian Research and Innovation Agency (ARIS) grant P2-0442.}

\newpage

\appendix
\section{Choosing the Wasserstein distance threshold}\label{app:sec:threshold}
\begin{figure}[htbp]
    \centering
    \includegraphics[width=0.9\linewidth]{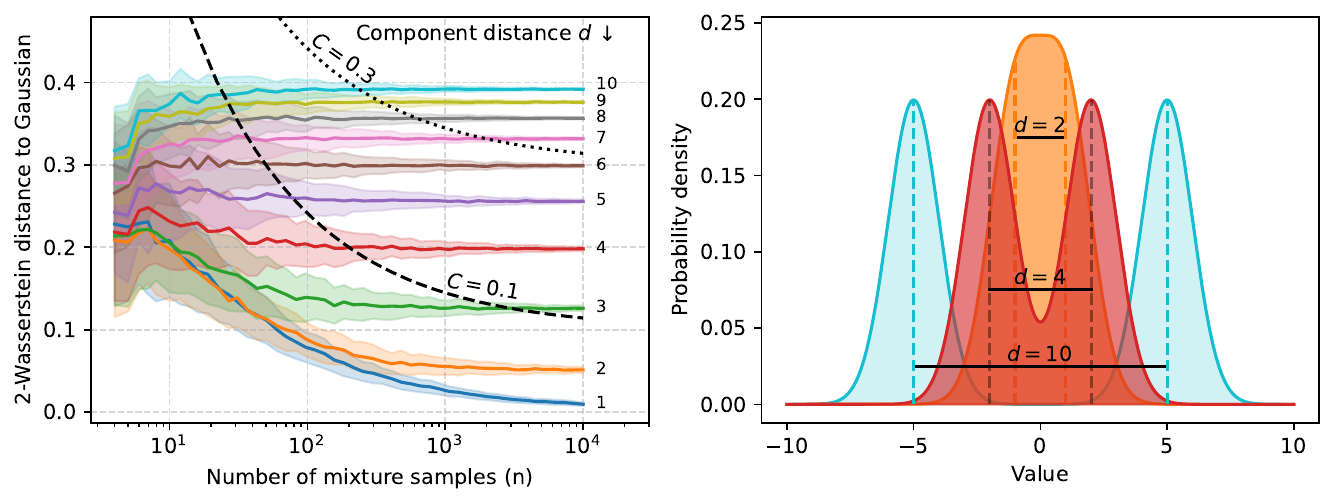}
    \caption{Left: approximate Gaussianity threshold as a function of sample size in relation to 2-Wasserstein distances on double Gaussian mixtures with component distance $d$. Colored lines denote the average $W_2$ to $N(0, 1)$ based on 150 sampled datasets of size $n$, shaded areas denote $\pm$ one standard deviation. Dashed and dotted lines denote thresholds $\tau$ given $C \in \{0.1, 0.3\}$. Right: illustrative examples for $d \in \{2,4,10\}$.}
    \label{fig:thresholds}
\end{figure}

When deciding on a value of $C$, we are essentially deciding which distribution shapes should be handled by a linearly-preconditioned MCMC kernel.
In Figure~\ref{fig:thresholds}, we visualize how the 2-Wasserstein distance to a Gaussian changes as we increase the distance between means in a one-dimensional double Gaussian mixture.
The mixture contains components $N(-d/2, 1)$ and $N(d/2, 1)$ with equal weights, where $d$ is the distance between component means.

Using $C=0.1$ allows the $d = 2$ case to be handled by a linearly-preconditioned kernel, which is reasonable due to the approximate Gaussianity of the target.
Doing the same for $d = 10$ is problematic because the distance between means is high, causing difficult-to-cross energy barriers during MCMC.
Applying the threshold to $d = 4$ is more nuanced, as such geometries can be handled by, e.g., linearly-preconditioned HMC or a similarly powerful kernel.
In such a case, adopting a more relaxed threshold like $C = 0.3$ treats the case as approximately Gaussian and thus only applies linear preconditioning.
Skipping such marginals reduces the complexity of the NF preconditioner, which could also lead to better preconditioning stability, i.e., a smaller chance of encountering an NF-induced singularity in preconditioned space.
Using $C = 0.1$ means modeling this marginal with an NF preconditioner. This could be beneficial if other dimensions inform the conditional distribution to a large degree, like the conditional dependence in Neal's funnel distribution.
It could also be beneficial if the inductive bias of the preconditioner allows for fast, high-quality fits given few training samples.
Reducing $C$ towards 0 would drive us towards full NF preconditioning. 
However, the suitability of such a choice is subject to NF limitations, particularly regarding training efficiency and numerical stability.

We also tested the accuracy of classifying approximately Gaussian dimensions using different values of $C$ in $\tau = C + \sqrt{2/n}$, where $n$ is the number of IID samples from a test distribution.
We used $n \in \{100, 1000, 10000 \}$.
We considered the following test distributions as approximately Gaussian:
\begin{itemize}[noitemsep, nolistsep]
    \item $N(0, 1)$,
    \item $N(8, 2^2)$,
    \item Equally-weighted mixture of $N(0.15, 1)$ and $N(-0.15, 1)$.
\end{itemize}
We considered the following test distributions as not approximately Gaussian: 
\begin{itemize}[noitemsep, nolistsep]
    \item Equally-weighted mixture of $N(8, 2^2)$ and $N(-8, 1)$,
    \item Equally-weighted mixture of $N(3, 1)$ and $N(-3, 1)$,
    \item Student's $t$ distribution with 1.5 degrees of freedom,
    \item $\mathrm{Cauchy}(1.5, 1.5)$,
    \item $\mathrm{Gamma}(1.5, 1.5)$ with the shape-rate parametrization,
    \item Conditional Gaussian $x | y \sim N(0, \exp(y))$ with $y \sim N(0, 3^2)$,
    \item Conditional Gaussian $x | y \sim N((3/100) y^2 - 3, 1)$ with $y \sim N(0, 10^2)$.
\end{itemize}
Using $C = 0.1$ correctly classified each distribution into its corresponding class.
Using $C = 0.3$ falsely classified the Gamma distribution and the two mixtures in the second group as approximately Gaussian. As discussed before, this is not necessarily problematic within preconditioned MCMC.

\section{Preconditioned Metropolis-Hastings warmup strategies}\label{app:sec:warmup}
The original NF preconditioning formulation by~\cite{hoffman_neutralizing_2019}, termed NeuTra MCMC, involves training the preconditioner once using stochastic variational inference (SVI) and then tuning other kernel parameters.
\cite{grenioux_sampling_2023a} found that such preconditioned HMC obtains a higher log posterior predictive distribution score than HMC on a sparse logistic regression analysis, which aligns with the findings by~\cite{hoffman_neutralizing_2019}.
However, an empirical analysis by~\cite{nabergoj_empirical_2024} recently found preconditioned HMC and preconditioned random-walk MH to obtain worse second moment estimates than their linearly-preconditioned counterparts, unless additionally using global NF proposals, revealing limitations of the original approach.
The problem could arise from only training the preconditioner once, as well as disregarding all observed chain states that could improve the fit.

Several MCMC samplers utilize chain states as training data for the NF.
When using NFs solely as global proposal kernels within adaptive MCMC,~\cite{gabrie_efficient_2021} and~\cite{gabrie_adaptive_2022} propose a single gradient descent update on NF parameters during each step using current states in a parallel-chain sampler.
\cite{samsonov_localglobal_2022} extend this idea for global NF proposals with a single gradient descent update every $K > 0$ steps using all chain states over $K$ steps.
Within sequential Monte Carlo (SMC), \cite{karamanis_accelerating_2022} use a similar approach for NF-preconditioned random-walk Metropolis transitions while fixing the proposal scale to the optimal $2.38 / \sqrt{D}$ for $D$-dimensional Gaussian targets, which assumes perfect NF preconditioning.
Alternatively, \cite{arbel_annealed_2021} and~\cite{matthews_continual_2022} use the HMC kernel for Metropolis-Hastings transitions and manually tune its step size using preliminary runs of SMC.

While these samplers utilize MCMC draws as NF training data, we specifically note that the step size is either unaffected (in local-global samplers) or is set manually (in sequential samplers).
Several Bayesian inference packages combine linear preconditioner tuning and step size adaptation~\citep{carpenter_stan_2017, bingham_pyro_2019, phan_composable_2019a, abril-pla_pymc_2023a, cabezas_blackjax_2024}.
The general warmup strategy consists of several cycles, each lasting a number of MCMC steps.
The preconditioner is updated at the end of each cycle using the observed chain states from within that cycle.
The step size is tuned while the preconditioner remains relatively stable, so the latent space does not change significantly.
We draw on these ideas when designing our warmup procedure for preconditioned Metropolis-Hastings.
We note that our method shares similarities with recent experimental developments in Nutpie~\citep{seyboldt_nutpie_2025}, an emerging general-purpose Bayesian inference library that implements the preconditioned NUTS using the RNVP architecture.

\subsection{Utilized warmup strategy}
We propose an MCMC warmup phase, where the preconditioner is trained in cycles, i.e., at regular intervals using data from MCMC exploration.
We outline the warmup procedure in Algorithm~\ref{alg:warmup}.
We note that the algorithm can be modified for NUTS by always accepting the transitions.

\begin{algorithm}[htbp]
\caption{Preconditioned Metropolis-Hastings warmup}
\begin{algorithmic}[1]
\Require Metropolis-Hastings kernel $K$ with invariant distribution $\pi$, initial latent states $\{z_0^{(\ell)}\}_{\ell=1}^k$ for $k \geq 1$ chains, per-chain step sizes $\{\epsilon^{(\ell)}\}_{\ell=1}^k$, preconditioner $P_\theta$, number of cycles $m \geq 1$, number of Metropolis-Hastings steps $n \geq 1$ within each cycle, maximum number of training samples $r  \geq 1$, target acceptance rate $\alpha^* \in (0, 1)$.
\Statex
\State Initialize reservoir $R$ with size $r$.
\For{$j = 1$ to $m$}
    \For{$i = 1$ to $n$}
        \State \texttt{/* Classic Metropolis-Hastings proposal routine */}
        \State Propose states $z_{i}^{(\ell)\prime} \sim K_\phi(z_{i-1}^{(\ell)}, \cdot)$ for $\ell=1, \dots, k$.
        \State Compute acceptance ratios $\{\alpha^{(\ell)}\}_{\ell=1}^k$.
        \State Generate thresholds $\{u^{(\ell)}\}_{\ell=1}^k$ with $u^{(\ell)}\sim_{\mathrm{IID}}\mathrm{Uniform}(0, 1)$.
        \Statex
        \State \texttt{/* Independently process chains \hfill */}
        \For{$\ell = 1$ to $k$}
            \State \texttt{/* Update state and store acceptance flag $a^{(\ell)}$ \hfill */}
            \If{$\alpha^{(\ell)} > u^{(\ell)}$}
                \State $z_i^{(\ell)} \gets z_i^{(\ell)\prime}$ and $a^{(\ell)} \gets 1$.
            \Else
                \State $z_i^{(\ell)} \gets z_{i-1}^{(\ell)}$ and $a^{(\ell)} \gets 0$.
            \EndIf

            \If{$i < n/2$}
                \State \texttt{/* Update step size in first half of cycle \hfill */}
                \State Update $\epsilon^{(\ell)}$ with a dual averaging step using error $\alpha^* - a^{(\ell)}$.
            \Else
                \State \texttt{/* Acquire training data with fixed step sizes \hfill */}
                \State Transform $x_i^{(\ell)} \gets P_\theta^{-1}(z_i^{(\ell)})$ and add $x_i^{(\ell)}$ to $R$.
            \EndIf
        \EndFor
    \EndFor
    \Statex
    \State \texttt{/* Avoid conflict between $P_\theta$ and step size on last cycle \hfill */}
    \If{$j < m$}
        \State \texttt{/* Train with subsampled data from all cycles so far \hfill */}
        \State Fit $P_\theta$ to target samples in $R$.
    \EndIf
\EndFor
\end{algorithmic}
\label{alg:warmup}
\end{algorithm}

\subsubsection{Maintaining IID training samples with limited memory}
The training data acquired through warmup cycles is added to a reservoir~\citep{vitter_random_1985}, which reduces memory requirements when dealing with long chains or many parallel chains.
This is accomplished by replacing a reservoir sample at index $i$ with the new sample when the reservoir is full, where $i \sim U(\{1, \dots, r\})$.
In the ideal case where all training data is IID from distribution $\pi$, reservoir samples are also IID according to $\pi$ in every MCMC step.
While this case is not satisfied for MCMC draws, the distribution is still largely retained, assuming chains mix fairly well in the second half of each cycle.

Using a reservoir thus ensures that the preconditioner has access to data from early cycles.
If chains sample inefficiently in early cycles, the training data is still largely replaced by samples from later cycles, and a robustly trained preconditioner can maintain a quality fit.
When dealing with highly autocorrelated samples, the reservoir's subsampling mechanism ensures that the training dataset is representative of the target distributions and not biased to highly similar samples from only the latest cycle.
This is also important for our proposed method, which rebuilds the conditional NF after every cycle.
In our implementation, we limit the reservoir size to 15000 samples to work within the memory constraints of our cluster when executing many parallel experiments. We did not observe significant differences when reducing the reservoir size to 5000.
We run 5 warmup cycles in all experiments, each cycle having 1000 iterations.
We found that increasing the number of cycles or iterations further did not significantly affect results.

\subsubsection{Tuning the step size}
In the first part of each cycle, we use Nesterov's dual averaging~\citep{hoffman_nouturn_2011} to tune the kernel step size according to acceptance data.
Importantly, we apply dual averaging for each chain individually, which allows them to adapt to their local geometry and find the typical set more easily.
In our experiments, we observed this to be crucial to avoid chains getting stuck, as only using a single step size often led to near-zero acceptance rates for some chains.
This somewhat inflated the number of non-representative training samples and worsened preconditioner fits.
We use hyperparameters $\gamma = 0.05, \kappa = 0.75, t_0 = 10$ as suggested default values~\citep{hoffman_nouturn_2011}.
We use an initial step size of 0.01, as initial values of 0.1 and 0.001 did not yield noticeably different results.

\subsubsection{Training the preconditioner}
In the second part of each cycle, we maintain a fixed step size for each individual chain and run MCMC.
The latent samples observed during this part are transformed to the target space via the inverse preconditioner transform and then added to the reservoir. 
The inverse transform can be performed after ensuring the sample will actually be placed into the reservoir, which can somewhat improve implementations with computationally demanding preconditioner inverses.
At the end of the cycle, the preconditioner parameters are optimized to fit the training data.
This involves estimating the training mean and covariance for linear preconditioners, or applying stochastic optimization for NFs.
In our implementation, we optimize NF parameters with AdamW using a learning rate of 0.001 for 3500 epochs.
This choice of learning rate allowed NFs to reach a stable minimum.
Increasing the number of epochs did not noticeably change the results.
For RNVP, the optimizer state is not reset between cycles, allowing optimization progress to be retained.

Simultaneously tuning the step size and the preconditioner could lead to instabilities.
We thus do not update the preconditioner at the end of the last cycle, ensuring that the preconditioned space does not change, and that the previously tuned step size remains valid.
The pseudocode in Algorithm~\ref{alg:warmup} is somewhat simplified, as the last cycle effectively only lasts for $n/2$ steps.
The first warmup cycle is always performed with an identity preconditioner, as there is no training data yet.
This functions as a burn-in phase where chains find the approximate region of the typical set.
We use a diagonal preconditioner in the second cycle for two reasons.
First, the training data from the first cycle can have a fairly simple, approximately Gaussian geometry, which can be modeled without an NF.
Second, sampling is somewhat faster using a diagonal preconditioner rather than an NF, as it avoids more involved NF transformations during kernel transitions.

\vskip 0.2in
\bibliography{bibliography}

\end{document}